\documentclass[10pt, a4paper]{article}

\usepackage[]{lrec-coling2024} % this is the new style

\newcommand{\absinth}{\textsc{absinth}} 
\usepackage{booktabs,array} % tables
\usepackage{comment}
\usepackage{pgfplots}
\pgfplotsset{compat=1.18} 
\usepackage{xcolor}

\usepackage{fdsymbol}
\title{German also Hallucinates! Inconsistency Detection in News Summaries with the Absinth Dataset}

\name{Laura Mascarell$^{\dag*}\thanks{*Equal contribution}$, Ribin Chalumattu$^{\dag*}$, Annette Rios$^{\S}$} 

\address{$^{\dag}$ETH Zurich, $^{\S}$University of Zurich\\
         lmascarell@inf.ethz.ch, cribin@inf.ethz.ch, arios@ifi.uzh.ch}

\abstract{The advent of Large Language Models (LLMs) has led to remarkable progress on a wide range of natural language processing tasks. Despite the advances, these large-sized models still suffer from hallucinating information in their output, which poses a major issue in automatic text summarization, as we must guarantee that the generated summary is consistent with the content of the source document. Previous research addresses the challenging task of detecting hallucinations in the output (i.e.\ inconsistency detection) in order to evaluate the faithfulness of the generated summaries. However, these works primarily focus on English and recent multilingual approaches lack German data. This work presents \absinth, a manually annotated dataset for hallucination detection in German news summarization and explores the capabilities of novel open-source LLMs on this task in both fine-tuning and in-context learning settings. We open-source and release the \absinth~dataset to foster further research on hallucination detection in German.
 \\ \newline \Keywords{Summarization, Natural Language Generation, Evaluation Methodologies, Corpora} }
 
\begin{document}

\maketitleabstract

\section{Introduction}
The field of natural language processing is currently undergoing a paradigm shift towards the use of Large Language Models (LLMs), showing a performance leap against the state-of-the-art pre-trained language models such as GPT-2~\citep{radford2019language} by increasing their parameter scale \citep{zhao2023survey}. Despite the emerging abilities of these LLMs, they are still prone to fabricate information, that is, to hallucinate. In particular for text summarization, there is no guarantee that the information in the generated summary is faithful to the source document~\citep{tam-etal-2023-evaluating}.

Most of the research on inconsistency detection in summarization is focused on English, relying on annotated data that is not available in other languages~\citep{goyal-durrett-2021-annotating,kryscinski-etal-2020-evaluating,durmus-etal-2020-feqa}. Recently, \citet{qiu2023detecting} and \citet{gekhman2023trueteacher} propose multilingual approaches and evaluate them on the XLSum~\citep{hasan-etal-2021-xl} and mFace~\citep{aharoni2022mface} datasets, respectively. Even though these datasets comprise 44 languages, they do not include German, making it infeasible to assess inconsistency detection in this language.

It is important to highlight that there is not yet a consensus in the research community on the appropriate level of granularity for tackling this task. For the sake of simplicity, some existing benchmarks provide overall summary-level annotations of faithfulness~\citep{li2023halueval,clark2023seahorse,aharoni2022mface}, thus making it challenging to pinpoint where the hallucination occurs. Furthermore, all hallucinations often fall under the same category. \citet{maynez-etal-2020-faithfulness} distinguish between intrinsic and extrinsic hallucinations, as those that are counterfactual and add information to the source, respectively (see Table~\ref{tab:hallucination-examples}), allowing for a more fine-grained approach to hallucination detection. 

%data comes from...
In this paper, we present \absinth,
%\footnote{The herbal spirit is said to cause hallucination.}
the first summarization dataset that is manually annotated for inconsistency detection in German.\footnote{\href{https://github.com/mediatechnologycenter/Absinth}{\absinth~GitHub repository}} \absinth~ consists of 4,314 summary sentence-level annotations that differentiate between intrinsic and extrinsic hallucinations. Additionally, the dataset comprises the outputs of multiple summarization models, ranging from the state-of-the-art pre-trained language models for German summarization to the latest prompt-based LLMs such as GPT-4~\citep{openai2023gpt4} and the open-source LLama 2~\citep{touvron2023llama}.~Finally, we assess the ability of recent open-source LLMs at detecting hallucination using our data and experiment with both fine-tuning and in-context learning to adapt the models to our task. We compare their performance with conventional transformer models such as mBERT~\cite{devlin-etal-2019-bert} and XLM-RoBERTa~\cite{conneau-etal-2020-unsupervised} fine-tuned on the task. Our results show that mBERT achieves the best overall performance, whereas there is room for improvement with LLMs.
%Our experimental results show the strong capabilities of in-context learning with only a few examples. 

%We hope that this work fosters further research on hallucination detection in German.
%We release \absinth~and hope that this work fosters further research on hallucination detection in German.

%To the best of our knowledge, we are the first to tackle inconsistency detection in German summarization, and provide a dataset for this purpose providing a dataset that 

%a major limitation is their inability to remain factually consistent with the respective input document. 

%% We then show the capabilities of 
\begin{table}
    \small
    \centering
    \begin{tabular}{l}
    %\hline 
    \toprule
    %\textbf{Source} \\ 
    %\midrule
    \textbf{Source}: Prof.\ Park awarded Nobel Prize in Physics.\\
    %\midrule
    %\bottomrule
    %\textbf{Summary sentences}\\
    \midrule
    \{\textbf{F}\} Nobel Physics Prize goes to Prof.\ Park.\\\midrule
    \{\textbf{I}\} Prof.\ Park awarded Nobel Prize in \textbf{\textcolor{violet}{Economics}}.\\\midrule
    \{\textbf{E}\} Prof.\ Park (\textbf{\textcolor{violet}{58}}) awarded Nobel Prize in Physics.\\
    \bottomrule
    \end{tabular}
    \caption{\label{tab:hallucination-examples} Examples faithful to the source (F), containing intrinsic (I), or extrinsic hallucinations (E).}
\end{table}

\section{The \absinth~Dataset}
\absinth~is a dataset of German news articles and their generated summaries that is manually annotated for hallucination detection. In particular, \absinth~consists of 4,314 article-summary sentence pairs with the associated label \textit{Faithful}, \textit{Intrinsic}, or \textit{Extrinsic} hallucination. In this section, we describe the construction of the dataset~(Section~\ref{sec:dataset_construction}), the annotation task (Section~\ref{sec:annotation}), and the final steps to build the dataset (Section~\ref{sec:final_dataset}).
%The articles come from the \textit{20Minuten} dataset~\cite{rios-etal-2021-new} and the summaries are generated using multiple summarization systems.
% cleaning of the data

% hyperparameters
\subsection{Dataset Construction}
\label{sec:dataset_construction}

Our hallucination dataset comprises a random sample of 200 articles from the \textit{20Minuten}~\citeplanguageresource{20Minuten} test set split\footnote{\url{https://github.com/ZurichNLP/20Minuten/tree/main/SwissText_2023}} and seven summaries per article that we generate using different models and approaches. These models include the multilingual transformer-based models mBART~\cite{liu-etal-2020-multilingual-denoising} and mLongT5~\cite{uthus2023mlongt5} fine-tuned on the \textit{20Minuten} training data. While mBART has been widely used for German summarization~\cite{liu-etal-2020-multilingual-denoising}, mLongT5 has been recently introduced to handle long text inputs.
%The gradient accumulation steps is 16, resulting in a total effective batch size of 32.

Furthermore, we consider the latest prompt-based LLMs, namely GPT-4~\cite{openai2023gpt4} and the open-source Llama~2 models~\cite{touvron2023llama}. Within the Llama~2 family, we employ Stable Beluga 2, a Llama 2 model with 70b parameters fine-tuned on an Orca style dataset~\cite{mukherjee2023orca}, along with a smaller Llama 2 model with 7b parameters that we fine-tune on \textit{20Minuten}.

Finally, we employ GPT-4 to generate additional hallucinated instances. To ensure that they are not straightforward to identify, we enforce intrinsic and extrinsic hallucinations that adhere to the context of the source article. We therefore provide both article and reference summary and design a prompt for each hallucination type as follows: To generate intrinsic hallucinations, our prompt instructs the model to subtly alter the reference summary such that each sentence in the summary becomes counterfactual to the article.~In contrast, our prompt to generate extrinsic hallucinations instructs the model to add information in the reference summary that is not present in the article without deviating from the article topic (see prompts in Appendix~\ref{sec:appendix_prompts}).

%\begin{table}
%\small
%\centering
%\begin{tabular}{lrrrrrr}
%%\hline 
%\toprule
%\textbf{Model} & \textbf{R1}$\uparrow$ &  \textbf{R2}$\uparrow$ & \textbf{RL}$\uparrow$ & \textbf{$\rho$}$\downarrow$ & \textbf{snt} & \textbf{sum}\\
%\midrule
%mBart & 32.7 & 12.7 & 23.1 & 5.4 & 12 & 42\\
%mLongT5 & 33.5 & 13.8 & 23.9 & 8.3 & 13 & 43\\
%\midrule
%GPT-4 & 31.9 & 11.6 & 21.2 & 3.1 & 23 & 72\\
%GPT-4$_{ext}$ & 65.7 & 56.2 & 64.3 & 1.4 & 24 & 87\\
%GPT-4$_{int}$ & 81.2 & 67.7 & 80.5 & 1.6 & 13 & 45\\
%\midrule
%SBeluga2 & 33.9 & 12.7 & 22.5 & 3.5 & 20 & 53\\
%Llama2$_{ft}$ & 32.4 & 11.6 & 23.0 & 2.2 & 11 & 39 \\
%\bottomrule
%\end{tabular}
%\caption{\label{tab:dataset-info} Comparison of the summarization models in \absinth~evaluated on the \textit{20Minuten} test set. Higher values of the extractive fragment density $\rho$ indicate higher extractiveness \cite{grusky-etal-2018-newsroom}. \textit{snt} and \textit{sum} are the average token length of the generated sentences and summaries, respectively.}
%\end{table}

%%% Statistics of models
%%% sentence token length
%%% heatmap with the ROUGE differences?

%% split and sentence-level split
\subsection{Annotation Task}
\label{sec:annotation}

\begin{table}
\small
\centering
\begin{tabular}{lrrrrrr}
%\hline 
\toprule
\textbf{Model} & \textbf{FT} & \textbf{R1}$\uparrow$ & \textbf{RL}$\uparrow$ & \textbf{$\rho$}$\downarrow$ & \textbf{snt} & \textbf{sum}\\
\midrule
mBART & 20m & 32.7 & 23.1 & 5.4 & 12 & 42\\
mLongT5 & 20m & 33.5 & 23.9 & 8.3 & 13 & 43\\
\midrule
GPT-4 & - & 31.9 & 21.2 & 3.1 & 23 & 72\\
GPT-4$_{ext}$ & - & 65.7 & 64.3 & 1.4 & 24 & 87\\
GPT-4$_{int}$ & - & 81.2 & 80.5 & 1.6 & 13 & 45\\
\midrule
SBeluga2 & - & 33.9 & 22.5 & 3.5 & 20 & 53\\
Llama2$_{ft}$ & 20m & 32.4 &  23.0 & 2.2 & 11 & 39 \\
\bottomrule
\end{tabular}
\caption{\label{tab:dataset-info} Comparison of the summarization models in \absinth~evaluated on the \textit{20Minuten} test set in terms of rouge-1 and rouge-L scores. The high rouge scores of GPT-4$_{ext}$ and GPT-4$_{int}$ are due to applying the hallucination changes directly in the reference summary. The FT column indicates whether the model is fine-tuned on  \textit{20Minuten}. Higher values of the extractive fragment density $\rho$ indicate higher extractiveness \cite{grusky-etal-2018-newsroom}. \textit{snt} and \textit{sum} are the average token length of the generated sentences and summaries, respectively.}
\end{table}

We design a task to manually annotate our dataset for hallucination detection. More specifically, given an article $A$ and a sentence of a generated summary $s$, the task is to assess the consistency of $s$ with the content of the source article $A$. If $s$ is entirely consistent with $A$, it must be annotated as \textit{Faithful}. In contrast, if $s$ contains hallucinated information, we distinguish between 
hallucinations that are counterfactual to the content of the article $A$ (\textit{Intrinsic Hallucination}) and those that add information and, therefore, cannot be verified against $A$ (\textit{Extrinsic Hallucination}). Finally, we provide a fourth label to indicate that $s$ contains both intrinsic and extrinsic hallucinations. We then recruit a team of 12 native German speakers to annotate the data, such that every article-sentence summary pair is reviewed by three different annotators.
%We divided our hallucination dataset into four sets of articles and their corresponding summaries. We then recruited a team of 12 native German speakers and assigned three annotators to each set.

%\begin{itemize}
%    \item \textit{Faithful}: all the information in the summary sentence can be verified against the article.
%    \item \textit{Intrinsic Hallucination}: some information in the summary sentence is counterfactual to the content of the source article.
%    \item \textit{Extrinsic Hallucination}: the summary sentence contains additional information that cannot be verified against the article.
%    \item \textit{Intrinsic and Extrinsic}: the summary sentence contains intrinsic and extrinsic hallucinations.
%\end{itemize}

%We divided our hallucination dataset into four sets of articles and their corresponding summaries and assigned three crowd-workers to each group. Our gold annotations are equally distributed among the sets such that each crowd-worker annotates a total of 50 different articles. Finally, we randomly shuffled the articles order for each crowd-worker to avoid any biases. The annotation task takes eight hours, and they were asked to complete it throughout two consecutive days. 

To ensure that the annotators follow our annotation scheme, we continuously evaluate their performance on a gold standard that we annotated internally. These gold annotations are equally distributed among the sets such that each set comprises 50 different articles. Additionally, the articles and summaries are randomly shuffled for each human annotator to avoid biases. The annotation of a full set takes eight hours, and they were asked to complete it throughout two consecutive days.

Besides the continuous evaluation, we also implemented the following strategies to ensure high-quality annotations and high-inter annotator agreement: (a) in-person training and clear annotation guidelines; (b) the use of an intuitive annotation framework; and (d) a fair pay that aligns with the hourly wage of teaching assistants. Overall, we obtain a Fleiss'$\kappa$ \cite{fleiss1971mns} agreement of 0.81 when distinguishing between \textit{Faithful} or \textit{Hallucination} and 0.77 on the four labels, indicating a very high agreement. Previous work reports a lower $\kappa$ of 0.65 with three annotators on a similar annotation task~\cite{falke-etal-2019-ranking}, which confirms the effectiveness of our annotation strategy.

%a continuous evaluation of the crowd-workers performance against our gold annotations;

%Besides the annotation of a gold standard, our main strategies to ensure high-quality annotations and high-inter annotator agreement are: (a) in-person training and clear annotation guidelines, (b) use of an intuitive annotation framework, and (c) continuous evaluation of the annotators performance against our gold annotations.\footnote{The annotators are offered a fair pair corresponding to the hourly rate of...}

%according to the definition in Section~\ref{sec:label_definition}
%We recruited a team of 12 native German speakers to annotate the dataset. To ensure high-quality annotations and high-inter annotator agreement, we built a gold standard and applied the following strategies: (a) in-person training and clear annotation guidelines, (b) use of an intuitive annotation framework, and (c) continuous evaluation of the annotators performance against the gold standard.\footnote{The annotators are offered a fair pair corresponding to the hourly rate of...} The latter allows us to promptly identify those annotators that need further clarification on the task. 

% The completion of the task amounts to 8 hours of work and we paid the annoations with the standard hourly rate of 

\paragraph{Gold Standard} 
% authors of the paper
Three domain experts annotate a gold standard consisting of 25 random articles from our dataset and their corresponding generated summaries. Since each summary contains about three sentences, our gold standard comprises a total of 580 article-sentence summary pairs. The purpose of the gold standard is twofold: Firstly, to identify annotation challenges beforehand, and secondly, to promptly assist those annotators that need further clarification on the task. The Fleiss'$\kappa$ agreement on \textit{Faithful} or \textit{Hallucination} and all four labels are 0.86 and 0.90, respectively. The experts reached a consensus on the final label for the instances with disagreement, except for three ambiguous instances that are discarded. 

%\paragraph{Training and Annotation Setup} The annotation of a Gold standard allowed us to define annotation examples for challenging instances.    
%- Clear instructions: give examples each category
%- We setup training tutorials for first time users to train and provide feedback on the task.  

\paragraph{Intuitive Annotation Framework} To annotate our dataset, we use doccano~\citeplanguageresource{doccano}, an open-source crowd-sourcing text annotation tool, and adapt the code to our task (see Appendix~\ref{sec:appendix_doccano}).
The framework also allows annotators to add comments such that we can gather more information to inspect ambiguous cases. 

%which provides us with an intuitive framework for annotators and administrators, and adapted the implementation to our task (see Appendix~\ref{sec:appendix_doccano}). The framework allows crowd-workers to add comments such that we can obtain more information to inspect ambiguous cases. 

\paragraph{Continuous Evaluation} We randomly intersperse our gold standard in the annotation data and monitor the performance of each annotator on the gold annotations to provide them with clarifications if necessary. Furthermore, our dataset contains 121 duplicated summary sentences as a result from generating multiple summaries of the same article. We also use these duplicates to monitor their performance. Essentially, if an annotator uses a different label for a duplicate, the annotator is possibly performing the overall task incorrectly. Ultimately, we had to replace one of the annotators due to bad performance on the gold standard samples even when there was no ambiguity.
%Despite giving further clarification on the task, one of the crowd-workers performed poorly on gold annotations where there was no ambiguity. Therefore, we finally replaced the annotations with the judgments of an additional annotator. 

\subsection{Final Dataset}
\begin{table}
\small
\centering
\begin{tabular}{lrrr}
%\hline 
\toprule
\textbf{Split} & \textbf{Faithful} & \textbf{Extrinsic} & \textbf{Intrinsic}\\
\midrule
Train & 1,957 & 512 & 522 \\
Validation & 132 & 42 & 28 \\
Test Gold & 353 & 92 & 104 \\
Test Crowd & 351 & 112 & 100 \\
\bottomrule
\end{tabular}
\caption{\label{tab:class-distribution} Class distribution in our \absinth~dataset.}
\end{table}
\label{sec:final_dataset}

\begin{figure}
    \centering
    \begin{tikzpicture}
    \begin{axis}[
        xbar stacked,
        %legend style={
        %    at={(1.45,0.5)},
        %    anchor=east,
        %    draw=gray,
        %    cells={anchor=west},
        %    legend columns=1,
        %    font=\footnotesize
        %},
        legend style={
         legend columns=3,
         at={(xticklabel cs:0.5)},
         anchor=north,
         draw=gray,
         yshift=-4pt % Adjust the yshift value for more space
        },
        ytick=data,
        axis y line*=none,
        axis x line*=bottom,
        tick label style={font=\footnotesize},
        legend style={font=\footnotesize},
        label style={font=\footnotesize},
        xtick={0,150,300,450,600,750},
        width=.47\textwidth,
        bar width=3.7mm,
        yticklabels={Llama2$_{ft}$,SBeluga2,GPT-4$_{int}$,GPT-4$_{ext}$,GPT-4,mLongT5, mBART},
        xmin=0,
        xmax=650,
        area legend,
        y=4.5mm,
        enlarge y limits={abs=0.625},
        %nodes near coords,
        %every node near coord/.append style={
        %    anchor=west,
        %    font=\footnotesize,
        %    /pgf/number format/precision=1
        %}
    ]
    \addplot[draw=none,fill=blue!30] coordinates {(534,0) (484,1) (87,2) (47,3) (574,4) (537,5) (530,6)};
    \addplot[draw=none,fill=yellow!60] coordinates {(59,0) (16,1) (541,2) (12,3) (7,4) (39,5) (80,6)};
    \addplot[draw=none,fill=orange!50] coordinates {(52,0) (2,1) (44,2) (638,3) (6,4) (7,5) (9,6)};
    \legend{Faithful, Intrinsic, Extrinsic}
    \end{axis}
    \end{tikzpicture}
    \caption{Class distribution for each summarization model in \absinth. The largest models GPT-4 and Stable Beluga 2 generate the least hallucinations. Since summaries are of different sentence length, the total of instances varies among models.}
    \label{fig:class-distribution}
\end{figure}

\begin{table*}[ht]
\small
\centering
\begin{tabular}{llrrrrr}
\toprule
%\hline 
 \textbf{Model} & \textbf{Setting} & \textbf{F$_1$ macro} & \textbf{F$_1$ Faithful} & \textbf{F$_1$ Intrinsic} & \textbf{F$_1$ Extrinsic} & \textbf{BACC} \\
\midrule
Llama2 7b & zero-shot & 0.265 & 0.776 & 0.019 & 0.0 & 0.335\\
Llama2 7b & few-shot~(3) & 0.226 & 0.318 & \textit{0.308} & 0.052 & \textit{0.344}\\ %\hline
%\midrule
Llama2 13b & zero-shot & 0.258 &0.774 &0.0 & 0.0 & 0.332\\
Llama2 13b & few-shot~(3) & \textit{0.280} & 0.290 & \textit{0.315} & \textit{0.237} & \textit{0.375}\\
\midrule
LeoLM-mistral 7b & zero-shot & 0.143 & 0.077 &	0.054 & 0.299 & 0.327\\
LeoLM-mistral 7b & few-shot~(3) & \textit{0.281} & \textit{0.415} & \textit{0.103} & \textit{0.326} & \textit{0.385}\\
\midrule
LeoLM 7b & zero-shot &	0.274 & 0.467 & 0.326 & 0.028 & 0.377\\
LeoLM 7b & few-shot~(3) & 0.103 & 0.0 & 0.0 &\textit{0.310} & 0.333\\%\hline
LeoLM 13b	& zero-shot & 0.258 &	0.773 & 0.0 & 0.0 & 0.331\\
LeoLM 13b & few-shot~(3) &  \textit{0.372} & 0.554	& \textit{0.241} & {0.321} & {0.419}\\
%LeoLM 13b & $\uparrow$ fine-tuning~(3) & 0.386	& 0.631	& 0.238 & 0.288 & 0.405 \\
LeoLM 13b & fine-tuning &  \textit{0.483} & \textbf{0.886} & {0.029}	& \textit{0.533} & \textit{0.530}\\
\midrule
mBERT & fine-tuning & \textbf{0.740} & \textbf{0.882} & \textbf{0.564} & \textbf{0.780} & \textbf{0.732}\\
XLM-RoBERTa& fine-tuning & 0.642 & 0.861 & 0.352 & 0.714 & 0.624\\
\bottomrule
\end{tabular}
\caption{\label{tab:f1-scores-results} Macro-averaged F$_1$, class-wise F$_1$, and BACC scores averaged over three seeds in different settings\textemdash i.e.\ fine-tuning, zero-shot, and three few-shot prompting\textemdash on our inconsistency detection task. We highlight the improvements over the corresponding zero-shot. The overall best performance is in bold.}
\end{table*}

To build the final dataset, we discard 121 duplicates and 11 instances with the label \textit{Intrinsic and Extrinsic}. We then assign the label with the majority vote to the rest.
Figure~\ref{fig:class-distribution} and Table~\ref{tab:class-distribution} show the distribution of the classes across the models and dataset splits, respectively.\footnote{Test gold class distribution after discarding three ambiguous instances, 22 duplicates, and six instances with the label \textit{Intrinsic and Extrinsic}.} The test split contains our gold annotations and 25 additional articles, where at least one annotator disagrees on multiple instances, under the assumption that those samples are more challenging to predict. Additionally, the dataset includes a set of 71 instances with no agreement. To distinguish these instances from the actual test set, we mark them as `full disagreement'. 

%\section{Inconsistency Detection with Open-source LLMs}
\section{Inconsistency Detection Task}
 Our multi-classification task consists on  predicting the faithfulness of a summary sentence to the source article (i.e.\ \textit{Faithul}, \textit{Intrinsic}, or \textit{Extrinsic} hallucination) according to the definition in Section~\ref{sec:annotation}. We then assess the performance of different open-source LLMs on the \absinth~test set in multiple settings, such as fine-tuning and in-context learning, where we extend the prompt with random examples on each label from the \absinth~training data. 

\subsection{Models Selection}
%We adopt models from the open-source Llama 2 family~\citep{touvron2023llama}, which shows high performance on a variety of tasks.\footnote{\href{https://huggingface.co/spaces/HuggingFaceH4/open_llm_leaderboard}{Open LLM Leaderboard}} Specifically, we experiment with base Llama 2 with 7B and 13B parameters.\footnote{\href{https://huggingface.co/NousResearch/Llama-2-13b-hf}{NousResearch/Llama-2}} Additionally, we consider LeoLM 7B and 13B models,\footnote{\href{https://huggingface.co/LeoLM/leo-hessianai-13b}{LeoLM/leo-hessianai}} which adapt Llama 2 to German through continued pretraining on German data, and Mistral 7b,\footnote{\href{https://huggingface.co/LeoLM/leo-mistral-hessianai-7b}{LeoLM/leo-mistral-hessianai-7b}} which outperforms Llama 2 on multiple benchmarks~\citep{jiang2023mistral}.
We adopt a wide range of models from the conventional mBERT and XLM-RoBERTa to a variety of open-source LLMs.~Specifically, we consider the Llama 2 family~\citep{touvron2023llama}, which shows high performance on different tasks,\footnote{\href{https://huggingface.co/spaces/HuggingFaceH4/open_llm_leaderboard}{Open LLM Leaderboard}} and experiment with base Llama 2 with 7b and 13b parameters. Additionally, we consider LeoLM 7b and 13b models, which adapt Llama 2 to German through continued pretraining on German data, and Mistral 7b, which outperforms Llama 2 on multiple benchmarks~\citep{jiang2023mistral}.

\subsection{Results}
Table~\ref{tab:f1-scores-results} shows the performance of the selected models in the zero-shot, three few-shot prompting, and prompt-based fine-tuning settings. We report macro-averaged F$_1$, class-wise F$_1$, and the balanced accuracy BACC scores\textemdash i.e.\ the average of accuracy scores from each class. The BACC scores are also adopted in the related work~\citep{kryscinski-etal-2020-evaluating,laban-etal-2022-summac} as they are not affected by the majority class~\citep{hanselowski-etal-2018-retrospective,tholke2023class}. %The scores are averaged over three seeds. 

We observe that the prompt-based LLMs improve the detection of intrinsic and extrinsic hallucination with the fine-tuning or the in-context learning setting, where models are prompted with three examples from our dataset. In particular, LeoLM 13b achieves the best performance, showing the benefits of further training on German data. However, LLMs exhibit a poor performance overall on this classification task.~In contrast, the conventional transformer models mBERT and XLM-RoBERTa perform remarkably well, with mBERT achieving the best performance across the three classes. These results are consistent with \citet{sun2023text}, where the authors claim that LLMs underperform fine-tuned models in text classification tasks.

Finally, we observe that the models are generally better at detecting extrinsic hallucinations than intrinsic hallucinations. The main difference between these types of hallucination is that the information labelled as extrinsic hallucination is not present in the source article. We suggest that in future work, LLMs could benefit from chain-of-thought prompting techniques that elicit reasoning in these models~\cite{wei2022chain} to improve their prediction of intrinsic hallucinations. 

%we observe that the smaller models struggle to learn in the in-context learning setting, where models are prompted with three examples from our dataset. In contrast, the larger 13b models show a performance increase across the three classes with both fine-tuning and few-shot prompting. These results are aligned with the recent findings in \citet{wei2023larger}, which prove that the ability of LLMs to learn in-context depends on the model scale, that is, the larger the better. 

\section{Related Work}
Previous work mostly focuses on the English language and implements inconsistency detection metrics in supervised and unsupervised settings~\citep{huang2021factual}. Whilst the former are trained on English datasets annotated for this task \citep{kryscinski-etal-2020-evaluating,goyal-durrett-2021-annotating}, the latter adopts existing models trained for Natural Language Inference (NLI) or question answering to detect inconsistencies in summaries \citep{falke-etal-2019-ranking, maynez-etal-2020-faithfulness, laban-etal-2022-summac, durmus-etal-2020-feqa}. Since these approaches rely on data and models that are limited to English, they cannot be directly applied to other languages. An exception is the XNLI dataset, the machine translated counterpart of the English NLI data. However, the dataset has been used in multilingual settings with unsatisfactory results~\citep{qiu2023detecting}.

More recent research implements multilingual approaches instead. \citet{qiu2023detecting} leverage machine translation to generate a multilingual labeled summarization dataset for inconsistency detection. To annotate the dataset, their approach combines the predictions of several inconsistency metrics for English. Similarly, \citet{gekhman2023trueteacher} annotate a multilingual training dataset using FLAN-PaLM 540b~\citep{chung2022scaling}, a LLM fine-tuned on the NLI task. Both approaches use their own synthetic dataset to fine-tune the multilingual pre-trained models BERT~\citep{devlin-etal-2019-bert} and T5~\citep{xue-etal-2021-mt5}, respectively, and evaluate their performance on mFace, a multilingual test set for factual consistency evaluation of abstractive summarization~\citep{aharoni2022mface}. Although mFace comprises 44 languages, it does not include German. Other approaches use ChatGPT\footnote{\url{https://openai.com/chatgpt}} to evaluate factual inconsistency \citep{luo2023chatgpt,li2023halueval}. However, the accuracy is only slightly above random chance. Additionally, \citet{aiyappa2023can} argue against using ChatGPT for evaluation, as we cannot guarantee that there is no training-test contamination.
%Previous works argue against using closed-source models for evaluation, as we cannot guarantee that there is no training-test contamination \citep{aiyappa2023can}. 
In contrast, our work compares the performance of recent open-source LLMs in both fine-tuning and in-context learning settings.

\section{Conclusion}
Due to the lack of German data for inconsistency detection, we present the \absinth~dataset, a collection of German news articles and their generated summaries that has been manually annotated for this task. The dataset provides summary sentence-level annotations that distinguish between hallucinations that are counterfactual to the article (intrinsic) and those that add information not present in the source (extrinsic), allowing for a more fine-grained approach to detecting hallucination.

We then evaluate the performance of novel open-source LLMs on this classification task using our data and experiment with different settings including few-shot prompting and prompt-based fine-tuning.~Whilst LLMs improve their performance with fine-tuning or three-shot prompting, they exhibit a poor overall performance.~Our results show that the conventional transformer model mBERT significantly outperforms the prompt-based models.  

%The experimental results show the strong capabilities of in-context learning with only three examples, and observed as in recent related work that the learning ability of LLMs improves with larger model scales. 

We expect this work to supplement and foster research on detecting hallucination that includes the German language, and we are excited to further explore this direction in future work.

\section{Ethics Statement}
We obtained the corresponding exemption determination (EK-2023-E-3) from the Ethics Commission of ETH Zurich university to perform the annotation task as it does not pose any risk for the annotators.
%We obtained the corresponding exemption determination from the Ethics Commission to perform the annotation task as it does not pose any risk for the annotators.
In addition, the annotations were anonymously collected and no conclusions can be drawn about any specific annotator.

The summaries in \absinth~are automatically generated, and we did not check them for problematic content such as hate speech or biases. Nevertheless, we do not anticipate further ethical issues besides those already identified in text generation~\cite{smiley-etal-2017-say,kreps_mccain_brundage_2022}.

\section{Limitations}
The articles used to create {\absinth} are part of the \textit{20Minuten} dataset \cite{rios-etal-2021-new}. We use the SwissText\_2023 test split \citeplanguageresource{20Minuten}, since this version has been filtered for duplicates and overlap with mc4 \citep{raffel2020mc4}, a multilingual dataset commonly used for pretraining LLMs.
However, since the dataset and the original news articles are available online, it is still possible that some of the newer LLM's might have seen these articles as part of their pre-training.
The annotated dataset is limited to news articles in Standard Swiss German from one particular news outlet, 20Minuten. The articles are in general relatively short and informal in style, but cover a wide range of topics. Models trained for faithfulness assessment on this data might not perform as well on longer, more complex texts.

\section{Acknowledgements}
This project is supported by Ringier, TX Group, NZZ, SRG, VSM, viscom, and the ETH Zurich Foundation. It is also funded by the Swiss Innovation Agency Innosuisse under grant agreement number PFFS-21-47.

%\subsection{Appendices}

%\nocite{*}
\section{Bibliographical References}\label{sec:reference}

\bibliographystyle{lrec-coling2024-natbib}
\bibliography{lrec-coling2024}

\begin{thebibliography}{2}
\expandafter\ifx\csname natexlab\endcsname\relax\def\natexlab#1{#1}\fi

\bibitem[{{Hiroki Nakayama et al.}(2018)}]{doccano}
{Hiroki Nakayama et al.} 2018.
\newblock \emph{doccano}.
\newblock distributed via github, 1.8.4.
\newblock PID \href{https://github.com/doccano/doccano}{https://github.com/doccano/doccano}.

\bibitem[{{Tannon Kew et al.}(2023)}]{20Minuten}
{Tannon Kew et al.} 2023.
\newblock \emph{20Minuten: A Multi-task News Summarisation Dataset for German}.
\newblock Department of Computational Linguistics, University of Zurich, distributed via github, 1.0.
\newblock PID \href{https://github.com/ZurichNLP/20Minuten}{https://github.com/ZurichNLP/20Minuten}.

\end{thebibliography}


\begin{thebibliography}{40}
\expandafter\ifx\csname natexlab\endcsname\relax\def\natexlab#1{#1}\fi

\bibitem[{Aharoni et~al.(2023)Aharoni, Narayan, Maynez, Herzig, Clark, and Lapata}]{aharoni2022mface}
Roee Aharoni, Shashi Narayan, Joshua Maynez, Jonathan Herzig, Elizabeth Clark, and Mirella Lapata. 2023.
\newblock \href {https://doi.org/10.18653/v1/2023.findings-acl.220} {Multilingual summarization with factual consistency evaluation}.
\newblock In \emph{Findings of the Association for Computational Linguistics: ACL 2023}, pages 3562--3591, Toronto, Canada. Association for Computational Linguistics.

\bibitem[{Aiyappa et~al.(2023)Aiyappa, An, Kwak, and Ahn}]{aiyappa2023can}
Rachith Aiyappa, Jisun An, Haewoon Kwak, and Yong-yeol Ahn. 2023.
\newblock \href {https://doi.org/10.18653/v1/2023.trustnlp-1.5} {Can we trust the evaluation on {C}hat{GPT}?}
\newblock In \emph{Proceedings of the 3rd Workshop on Trustworthy Natural Language Processing (TrustNLP 2023)}, pages 47--54, Toronto, Canada. Association for Computational Linguistics.

\bibitem[{Chung et~al.(2022)Chung, Hou, Longpre, Zoph, Tay, Fedus, Li, Wang, Dehghani, Brahma et~al.}]{chung2022scaling}
Hyung~Won Chung, Le~Hou, Shayne Longpre, Barret Zoph, Yi~Tay, William Fedus, Eric Li, Xuezhi Wang, Mostafa Dehghani, Siddhartha Brahma, et~al. 2022.
\newblock \href {https://doi.org/10.48550/arXiv.2210.11416} {Scaling instruction-finetuned language models}.
\newblock \emph{arXiv preprint arXiv:2210.11416}.

\bibitem[{Clark et~al.(2023)Clark, Rijhwani, Gehrmann, Maynez, Aharoni, Nikolaev, Sellam, Siddhant, Das, and Parikh}]{clark2023seahorse}
Elizabeth Clark, Shruti Rijhwani, Sebastian Gehrmann, Joshua Maynez, Roee Aharoni, Vitaly Nikolaev, Thibault Sellam, Aditya Siddhant, Dipanjan Das, and Ankur Parikh. 2023.
\newblock \href {https://doi.org/10.18653/v1/2023.emnlp-main.584} {{SEAHORSE}: A multilingual, multifaceted dataset for summarization evaluation}.
\newblock In \emph{Proceedings of the 2023 Conference on Empirical Methods in Natural Language Processing}, pages 9397--9413, Singapore. Association for Computational Linguistics.

\bibitem[{Conneau et~al.(2020)Conneau, Khandelwal, Goyal, Chaudhary, Wenzek, Guzm{\'a}n, Grave, Ott, Zettlemoyer, and Stoyanov}]{conneau-etal-2020-unsupervised}
Alexis Conneau, Kartikay Khandelwal, Naman Goyal, Vishrav Chaudhary, Guillaume Wenzek, Francisco Guzm{\'a}n, Edouard Grave, Myle Ott, Luke Zettlemoyer, and Veselin Stoyanov. 2020.
\newblock \href {https://doi.org/10.18653/v1/2020.acl-main.747} {Unsupervised cross-lingual representation learning at scale}.
\newblock In \emph{Proceedings of the 58th Annual Meeting of the Association for Computational Linguistics}, pages 8440--8451, Online. Association for Computational Linguistics.

\bibitem[{Dettmers et~al.(2024)Dettmers, Pagnoni, Holtzman, and Zettlemoyer}]{dettmers2023qlora}
Tim Dettmers, Artidoro Pagnoni, Ari Holtzman, and Luke Zettlemoyer. 2024.
\newblock \href {https://proceedings.neurips.cc/paper_files/paper/2023/hash/1feb87871436031bdc0f2beaa62a049b-Abstract-Conference.html} {Qlora: Efficient finetuning of quantized llms}.
\newblock \emph{Advances in Neural Information Processing Systems}, 36.

\bibitem[{Devlin et~al.(2019)Devlin, Chang, Lee, and Toutanova}]{devlin-etal-2019-bert}
Jacob Devlin, Ming-Wei Chang, Kenton Lee, and Kristina Toutanova. 2019.
\newblock \href {https://doi.org/10.18653/v1/N19-1423} {{BERT}: Pre-training of deep bidirectional transformers for language understanding}.
\newblock In \emph{Proceedings of the 2019 Conference of the North {A}merican Chapter of the Association for Computational Linguistics: Human Language Technologies, Volume 1 (Long and Short Papers)}, pages 4171--4186, Minneapolis, Minnesota. Association for Computational Linguistics.

\bibitem[{Durmus et~al.(2020)Durmus, He, and Diab}]{durmus-etal-2020-feqa}
Esin Durmus, He~He, and Mona Diab. 2020.
\newblock \href {https://doi.org/10.18653/v1/2020.acl-main.454} {{FEQA}: A question answering evaluation framework for faithfulness assessment in abstractive summarization}.
\newblock In \emph{Proceedings of the 58th Annual Meeting of the Association for Computational Linguistics}, pages 5055--5070, Online. Association for Computational Linguistics.

\bibitem[{Falke et~al.(2019)Falke, Ribeiro, Utama, Dagan, and Gurevych}]{falke-etal-2019-ranking}
Tobias Falke, Leonardo F.~R. Ribeiro, Prasetya~Ajie Utama, Ido Dagan, and Iryna Gurevych. 2019.
\newblock \href {https://doi.org/10.18653/v1/P19-1213} {Ranking generated summaries by correctness: An interesting but challenging application for natural language inference}.
\newblock In \emph{Proceedings of the 57th Annual Meeting of the Association for Computational Linguistics}, pages 2214--2220, Florence, Italy. Association for Computational Linguistics.

\bibitem[{Fleiss(1971)}]{fleiss1971mns}
J.L. Fleiss. 1971.
\newblock \href {https://doi.org/10.1037/h0031619} {Measuring nominal scale agreement among many raters}.
\newblock \emph{Psychological Bulletin}, 76(5):378--382.

\bibitem[{Gao et~al.(2023)Gao, Tow, Abbasi, Biderman, Black, DiPofi, Foster, Golding, Hsu, Le~Noac'h, Li, McDonell, Muennighoff, Ociepa, Phang, Reynolds, Schoelkopf, Skowron, Sutawika, Tang, Thite, Wang, Wang, and Zou}]{eval-harness}
Leo Gao, Jonathan Tow, Baber Abbasi, Stella Biderman, Sid Black, Anthony DiPofi, Charles Foster, Laurence Golding, Jeffrey Hsu, Alain Le~Noac'h, Haonan Li, Kyle McDonell, Niklas Muennighoff, Chris Ociepa, Jason Phang, Laria Reynolds, Hailey Schoelkopf, Aviya Skowron, Lintang Sutawika, Eric Tang, Anish Thite, Ben Wang, Kevin Wang, and Andy Zou. 2023.
\newblock \href {https://doi.org/10.5281/zenodo.10256836} {A framework for few-shot language model evaluation}.

\bibitem[{Gekhman et~al.(2023)Gekhman, Herzig, Aharoni, Elkind, and Szpektor}]{gekhman2023trueteacher}
Zorik Gekhman, Jonathan Herzig, Roee Aharoni, Chen Elkind, and Idan Szpektor. 2023.
\newblock \href {https://doi.org/10.18653/v1/2023.emnlp-main.127} {{T}rue{T}eacher: Learning factual consistency evaluation with large language models}.
\newblock In \emph{Proceedings of the 2023 Conference on Empirical Methods in Natural Language Processing}, pages 2053--2070, Singapore. Association for Computational Linguistics.

\bibitem[{Goyal and Durrett(2021)}]{goyal-durrett-2021-annotating}
Tanya Goyal and Greg Durrett. 2021.
\newblock \href {https://doi.org/10.18653/v1/2021.naacl-main.114} {Annotating and modeling fine-grained factuality in summarization}.
\newblock In \emph{Proceedings of the 2021 Conference of the North American Chapter of the Association for Computational Linguistics: Human Language Technologies}, pages 1449--1462, Online. Association for Computational Linguistics.

\bibitem[{Grusky et~al.(2018)Grusky, Naaman, and Artzi}]{grusky-etal-2018-newsroom}
Max Grusky, Mor Naaman, and Yoav Artzi. 2018.
\newblock \href {https://doi.org/10.18653/v1/N18-1065} {{N}ewsroom: A dataset of 1.3 million summaries with diverse extractive strategies}.
\newblock In \emph{Proceedings of the 2018 Conference of the North {A}merican Chapter of the Association for Computational Linguistics: Human Language Technologies, Volume 1 (Long Papers)}, pages 708--719, New Orleans, Louisiana. Association for Computational Linguistics.

\bibitem[{Hanselowski et~al.(2018)Hanselowski, PVS, Schiller, Caspelherr, Chaudhuri, Meyer, and Gurevych}]{hanselowski-etal-2018-retrospective}
Andreas Hanselowski, Avinesh PVS, Benjamin Schiller, Felix Caspelherr, Debanjan Chaudhuri, Christian~M. Meyer, and Iryna Gurevych. 2018.
\newblock \href {https://aclanthology.org/C18-1158} {A retrospective analysis of the fake news challenge stance-detection task}.
\newblock In \emph{Proceedings of the 27th International Conference on Computational Linguistics}, pages 1859--1874, Santa Fe, New Mexico, USA. Association for Computational Linguistics.

\bibitem[{Hasan et~al.(2021)Hasan, Bhattacharjee, Islam, Mubasshir, Li, Kang, Rahman, and Shahriyar}]{hasan-etal-2021-xl}
Tahmid Hasan, Abhik Bhattacharjee, Md.~Saiful Islam, Kazi Mubasshir, Yuan-Fang Li, Yong-Bin Kang, M.~Sohel Rahman, and Rifat Shahriyar. 2021.
\newblock \href {https://doi.org/10.18653/v1/2021.findings-acl.413} {{XL}-sum: Large-scale multilingual abstractive summarization for 44 languages}.
\newblock In \emph{Findings of the Association for Computational Linguistics: ACL-IJCNLP 2021}, pages 4693--4703, Online. Association for Computational Linguistics.

\bibitem[{Huang et~al.(2021)Huang, Feng, Feng, and Qin}]{huang2021factual}
Yichong Huang, Xiachong Feng, Xiaocheng Feng, and Bing Qin. 2021.
\newblock \href {https://doi.org/10.48550/arXiv.2104.14839} {The factual inconsistency problem in abstractive text summarization: A survey}.
\newblock \emph{arXiv preprint arXiv:2104.14839}.

\bibitem[{Jiang et~al.(2023)Jiang, Sablayrolles, Mensch, Bamford, Chaplot, Casas, Bressand, Lengyel, Lample, Saulnier et~al.}]{jiang2023mistral}
Albert~Q Jiang, Alexandre Sablayrolles, Arthur Mensch, Chris Bamford, Devendra~Singh Chaplot, Diego de~las Casas, Florian Bressand, Gianna Lengyel, Guillaume Lample, Lucile Saulnier, et~al. 2023.
\newblock \href {https://doi.org/10.48550/arXiv.2310.06825} {Mistral 7b}.
\newblock \emph{arXiv preprint arXiv:2310.06825}.

\bibitem[{Kreps et~al.(2022)Kreps, McCain, and Brundage}]{kreps_mccain_brundage_2022}
Sarah Kreps, R.~Miles McCain, and Miles Brundage. 2022.
\newblock \href {https://doi.org/10.1017/XPS.2020.37} {All the news that’s fit to fabricate: {AI}-generated text as a tool of media misinformation}.
\newblock \emph{Journal of Experimental Political Science}, 9(1):104–117.

\bibitem[{Kryscinski et~al.(2020)Kryscinski, McCann, Xiong, and Socher}]{kryscinski-etal-2020-evaluating}
Wojciech Kryscinski, Bryan McCann, Caiming Xiong, and Richard Socher. 2020.
\newblock \href {https://doi.org/10.18653/v1/2020.emnlp-main.750} {Evaluating the factual consistency of abstractive text summarization}.
\newblock In \emph{Proceedings of the 2020 Conference on Empirical Methods in Natural Language Processing (EMNLP)}, pages 9332--9346, Online. Association for Computational Linguistics.

\bibitem[{Laban et~al.(2022)Laban, Schnabel, Bennett, and Hearst}]{laban-etal-2022-summac}
Philippe Laban, Tobias Schnabel, Paul~N. Bennett, and Marti~A. Hearst. 2022.
\newblock \href {https://doi.org/10.1162/tacl_a_00453} {{S}umma{C}: Re-visiting {NLI}-based models for inconsistency detection in summarization}.
\newblock \emph{Transactions of the Association for Computational Linguistics}, 10:163--177.

\bibitem[{Li et~al.(2023)Li, Cheng, Zhao, Nie, and Wen}]{li2023halueval}
Junyi Li, Xiaoxue Cheng, Xin Zhao, Jian-Yun Nie, and Ji-Rong Wen. 2023.
\newblock \href {https://doi.org/10.18653/v1/2023.emnlp-main.397} {{H}alu{E}val: A large-scale hallucination evaluation benchmark for large language models}.
\newblock In \emph{Proceedings of the 2023 Conference on Empirical Methods in Natural Language Processing}, pages 6449--6464, Singapore. Association for Computational Linguistics.

\bibitem[{Liu et~al.(2020)Liu, Gu, Goyal, Li, Edunov, Ghazvininejad, Lewis, and Zettlemoyer}]{liu-etal-2020-multilingual-denoising}
Yinhan Liu, Jiatao Gu, Naman Goyal, Xian Li, Sergey Edunov, Marjan Ghazvininejad, Mike Lewis, and Luke Zettlemoyer. 2020.
\newblock \href {https://doi.org/10.1162/tacl_a_00343} {Multilingual denoising pre-training for neural machine translation}.
\newblock \emph{Transactions of the Association for Computational Linguistics}, 8:726--742.

\bibitem[{Luo et~al.(2023)Luo, Xie, and Ananiadou}]{luo2023chatgpt}
Zheheng Luo, Qianqian Xie, and Sophia Ananiadou. 2023.
\newblock \href {https://doi.org/10.48550/arXiv.2303.15621} {Chatgpt as a factual inconsistency evaluator for text summarization}.
\newblock \emph{arXiv preprint arXiv:2303.15621}.

\bibitem[{Maynez et~al.(2020)Maynez, Narayan, Bohnet, and McDonald}]{maynez-etal-2020-faithfulness}
Joshua Maynez, Shashi Narayan, Bernd Bohnet, and Ryan McDonald. 2020.
\newblock \href {https://doi.org/10.18653/v1/2020.acl-main.173} {On faithfulness and factuality in abstractive summarization}.
\newblock In \emph{Proceedings of the 58th Annual Meeting of the Association for Computational Linguistics}, pages 1906--1919, Online. Association for Computational Linguistics.

\bibitem[{Mukherjee et~al.(2023)Mukherjee, Mitra, Jawahar, Agarwal, Palangi, and Awadallah}]{mukherjee2023orca}
Subhabrata Mukherjee, Arindam Mitra, Ganesh Jawahar, Sahaj Agarwal, Hamid Palangi, and Ahmed Awadallah. 2023.
\newblock \href {https://doi.org/https://doi.org/10.48550/arXiv.2306.02707} {Orca: Progressive learning from complex explanation traces of gpt-4}.
\newblock \emph{arXiv preprint arXiv:2306.02707}.

\bibitem[{OpenAI(2023)}]{openai2023gpt4}
OpenAI. 2023.
\newblock Gpt-4 technical report.
\newblock \emph{arXiv preprint arXiv:2303.08774}.

\bibitem[{Qiu et~al.(2023)Qiu, Ziser, Korhonen, Ponti, and Cohen}]{qiu2023detecting}
Yifu Qiu, Yftah Ziser, Anna Korhonen, Edoardo~M Ponti, and Shay~B Cohen. 2023.
\newblock \href {https://doi.org/10.48550/arXiv.2305.13632"} {Detecting and mitigating hallucinations in multilingual summarisation}.
\newblock \emph{arXiv preprint arXiv:2305.13632}.

\bibitem[{Radford et~al.(2019)Radford, Wu, Child, Luan, Amodei, Sutskever et~al.}]{radford2019language}
Alec Radford, Jeffrey Wu, Rewon Child, David Luan, Dario Amodei, Ilya Sutskever, et~al. 2019.
\newblock Language models are unsupervised multitask learners.
\newblock \emph{OpenAI blog}, 1(8):9.

\bibitem[{Raffel et~al.(2020)Raffel, Shazeer, Roberts, Lee, Narang, Matena, Zhou, Li, and Liu}]{raffel2020mc4}
Colin Raffel, Noam Shazeer, Adam Roberts, Katherine Lee, Sharan Narang, Michael Matena, Yanqi Zhou, Wei Li, and Peter~J. Liu. 2020.
\newblock \href {http://jmlr.org/papers/v21/20-074.html} {Exploring the limits of transfer learning with a unified text-to-text transformer}.
\newblock \emph{Journal of Machine Learning Research}, 21(140):1--67.

\bibitem[{Rios et~al.(2021)Rios, Spring, Kew, Kostrzewa, S{\"a}uberli, M{\"u}ller, and Ebling}]{rios-etal-2021-new}
Annette Rios, Nicolas Spring, Tannon Kew, Marek Kostrzewa, Andreas S{\"a}uberli, Mathias M{\"u}ller, and Sarah Ebling. 2021.
\newblock \href {https://doi.org/10.18653/v1/2021.newsum-1.16} {A new dataset and efficient baselines for document-level text simplification in {G}erman}.
\newblock In \emph{Proceedings of the Third Workshop on New Frontiers in Summarization}, pages 152--161, Online and in Dominican Republic. Association for Computational Linguistics.

\bibitem[{Smiley et~al.(2017)Smiley, Schilder, Plachouras, and Leidner}]{smiley-etal-2017-say}
Charese Smiley, Frank Schilder, Vassilis Plachouras, and Jochen~L. Leidner. 2017.
\newblock \href {https://doi.org/10.18653/v1/W17-1613} {Say the right thing right: Ethics issues in natural language generation systems}.
\newblock In \emph{Proceedings of the First {ACL} Workshop on Ethics in Natural Language Processing}, pages 103--108, Valencia, Spain. Association for Computational Linguistics.

\bibitem[{Sun et~al.(2023)Sun, Li, Li, Wu, Guo, Zhang, and Wang}]{sun2023text}
Xiaofei Sun, Xiaoya Li, Jiwei Li, Fei Wu, Shangwei Guo, Tianwei Zhang, and Guoyin Wang. 2023.
\newblock \href {https://doi.org/https://doi.org/10.48550/arXiv.2305.08377} {Text classification via large language models}.
\newblock \emph{arXiv preprint arXiv:2305.08377}.

\bibitem[{Tam et~al.(2023)Tam, Mascarenhas, Zhang, Kwan, Bansal, and Raffel}]{tam-etal-2023-evaluating}
Derek Tam, Anisha Mascarenhas, Shiyue Zhang, Sarah Kwan, Mohit Bansal, and Colin Raffel. 2023.
\newblock \href {https://doi.org/10.18653/v1/2023.findings-acl.322} {Evaluating the factual consistency of large language models through news summarization}.
\newblock In \emph{Findings of the Association for Computational Linguistics: ACL 2023}, pages 5220--5255, Toronto, Canada. Association for Computational Linguistics.

\bibitem[{Th{\"o}lke et~al.(2023)Th{\"o}lke, Mantilla-Ramos, Abdelhedi, Maschke, Dehgan, Harel, Kemtur, Berrada, Sahraoui, Young et~al.}]{tholke2023class}
Philipp Th{\"o}lke, Yorguin-Jose Mantilla-Ramos, Hamza Abdelhedi, Charlotte Maschke, Arthur Dehgan, Yann Harel, Anirudha Kemtur, Loubna~Mekki Berrada, Myriam Sahraoui, Tammy Young, et~al. 2023.
\newblock \href {https://doi.org/https://doi.org/10.1016/j.neuroimage.2023.120253} {Class imbalance should not throw you off balance: Choosing the right classifiers and performance metrics for brain decoding with imbalanced data}.
\newblock \emph{NeuroImage}, 277.

\bibitem[{Touvron et~al.(2023)Touvron, Martin, Stone, Albert, Almahairi, Babaei, Bashlykov, Batra, Bhargava, Bhosale et~al.}]{touvron2023llama}
Hugo Touvron, Louis Martin, Kevin Stone, Peter Albert, Amjad Almahairi, Yasmine Babaei, Nikolay Bashlykov, Soumya Batra, Prajjwal Bhargava, Shruti Bhosale, et~al. 2023.
\newblock \href {https://doi.org/10.48550/arXiv.2307.09288} {Llama 2: Open foundation and fine-tuned chat models}.
\newblock \emph{arXiv preprint arXiv:2307.09288}.

\bibitem[{Uthus et~al.(2023)Uthus, Ontanon, Ainslie, and Guo}]{uthus2023mlongt5}
David Uthus, Santiago Ontanon, Joshua Ainslie, and Mandy Guo. 2023.
\newblock \href {https://doi.org/10.18653/v1/2023.findings-emnlp.628} {m{L}ong{T}5: A multilingual and efficient text-to-text transformer for longer sequences}.
\newblock In \emph{Findings of the Association for Computational Linguistics: EMNLP 2023}, pages 9380--9386, Singapore. Association for Computational Linguistics.

\bibitem[{Wei et~al.(2022)Wei, Wang, Schuurmans, Bosma, Ichter, Xia, Chi, Le, and Zhou}]{wei2022chain}
Jason Wei, Xuezhi Wang, Dale Schuurmans, Maarten Bosma, Brian Ichter, Fei Xia, Ed~Chi, Quoc~V Le, and Denny Zhou. 2022.
\newblock \href {https://proceedings.neurips.cc/paper_files/paper/2022/file/9d5609613524ecf4f15af0f7b31abca4-Paper-Conference.pdf} {Chain-of-thought prompting elicits reasoning in large language models}.
\newblock In \emph{Advances in Neural Information Processing Systems}, volume~35, pages 24824--24837. Advances in Neural Information Processing Systems.

\bibitem[{Xue et~al.(2021)Xue, Constant, Roberts, Kale, Al-Rfou, Siddhant, Barua, and Raffel}]{xue-etal-2021-mt5}
Linting Xue, Noah Constant, Adam Roberts, Mihir Kale, Rami Al-Rfou, Aditya Siddhant, Aditya Barua, and Colin Raffel. 2021.
\newblock \href {https://doi.org/10.18653/v1/2021.naacl-main.41} {m{T}5: A massively multilingual pre-trained text-to-text transformer}.
\newblock In \emph{Proceedings of the 2021 Conference of the North American Chapter of the Association for Computational Linguistics: Human Language Technologies}, pages 483--498, Online. Association for Computational Linguistics.

\bibitem[{Zhao et~al.(2023)Zhao, Zhou, Li, Tang, Wang, Hou, Min, Zhang, Zhang, Dong et~al.}]{zhao2023survey}
Wayne~Xin Zhao, Kun Zhou, Junyi Li, Tianyi Tang, Xiaolei Wang, Yupeng Hou, Yingqian Min, Beichen Zhang, Junjie Zhang, Zican Dong, et~al. 2023.
\newblock \href {https://doi.org/10.48550/arXiv.2303.18223} {A survey of large language models}.
\newblock \emph{arXiv preprint arXiv:2303.18223}.

\end{thebibliography}

\section{Language Resource References}
\label{lr:ref}
\bibliographystylelanguageresource{lrec-coling2024-natbib}
\bibliographylanguageresource{languageresource}

%\newpage
\appendix

\section{Annotation Framework}
\label{sec:appendix_doccano}
\begin{figure*}[ht]
    \centering
     \includegraphics[scale=0.25]{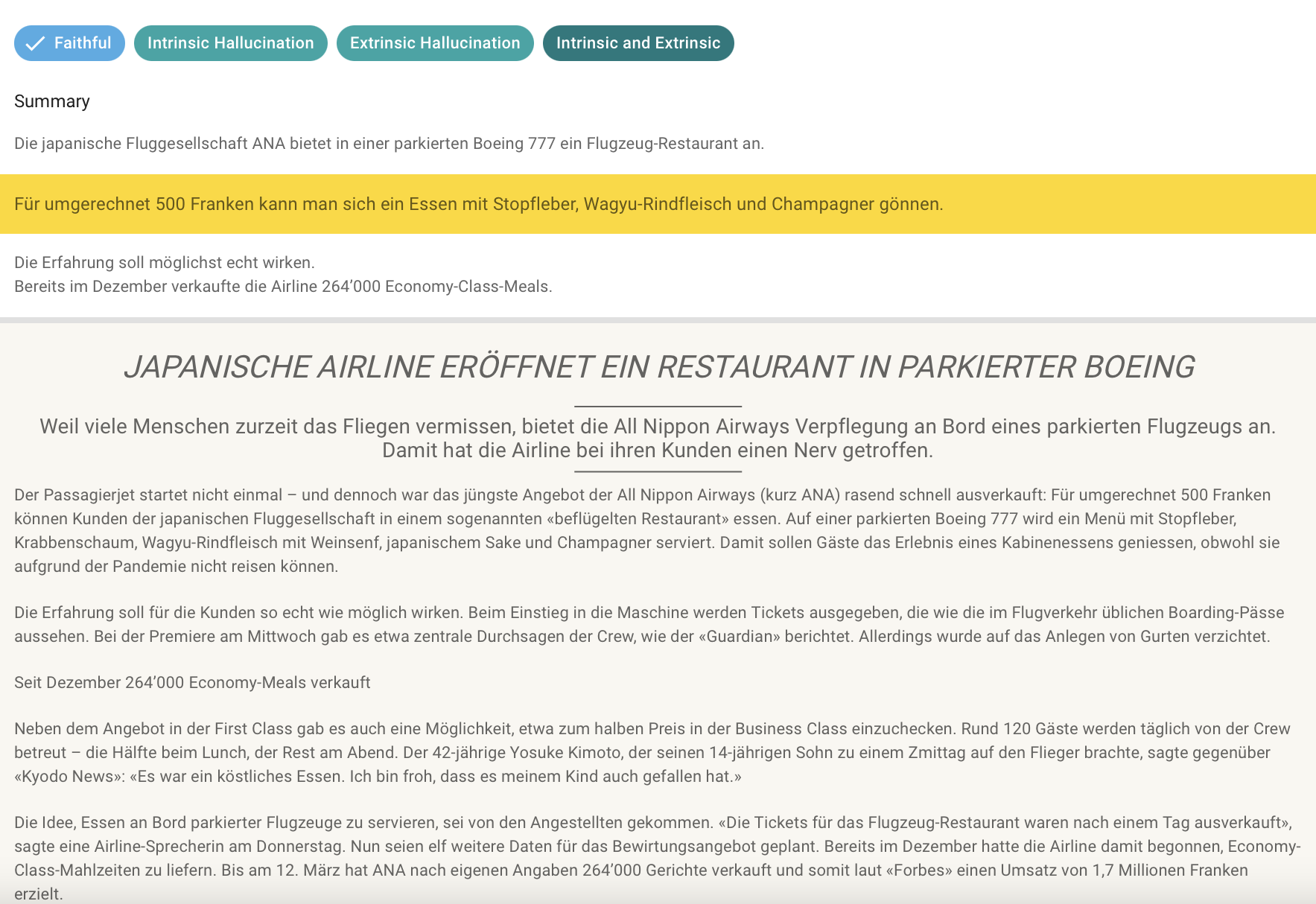}
     \caption{User interface of the annotation framework. We provide the article and all summary sentences. The interface highlights the summary sentence that is currently being reviewed.}
     \label{fig:appendix}
\end{figure*}

We extend doccano and implement a user interface to annotate article-summary sentence pairs. See an example in Figure~\ref{fig:appendix}.

\begin{table*}[ht]
\small
\centering
\begin{tabular}{lrrrrr}
\toprule
\textbf{Model} & \textbf{Training Set} & \textbf{Epochs} & \textbf{Learning Rate} & \textbf{Batch Size} & \textbf{Context Window} \\ \midrule
LLama 2 7b* & \textit{20Minuten} & 5 & $2e-4$ & 8 & 4,096 \\
mBart & \textit{20Minuten} & 10 & $3e-5$ & 32 & 1,024 \\
mLongT5 & \textit{20Minuten} & 10 & $3e-5$ & 32 & 2,048 \\
\midrule
LeoLM 13b* & \absinth & 1 & $2e-4$ & 8 & 4,096 \\
mBERT & \absinth & 5 & $2e-5$ & 32 & 512 \\
XLM-RoBERTa & \absinth & 5 & $2e-5$ & 32 & 512 \\
\bottomrule
\end{tabular}
\caption{Model fine-tuning details. The asterisk (*) indicates that the model is fine-tuned with QLoRA. }
\label{finetuning_details}
\end{table*}

\section{Prompts}
\label{sec:appendix_prompts}
Table~\ref{tab:prompts} lists all the prompts that we use in this work. We design these prompts using gpt-prompt-engineer.\footnote{\url{https://github.com/mshumer/gpt-prompt-engineer}}

\begin{table*}
\small
\centering
\begin{tabular}{p{15.4cm}l}
\textbf{GPT-4 Summarization}\\
\toprule 
Provide a concise, 3-sentence summary of the following news article. The summary MUST be written in German.\\
Article: \{article\}\\
\midrule
\end{tabular}

\begin{tabular}{p{15.4cm}l}
\\
\textbf{GPT-4 Intrinsic Hallucination}\\
\toprule 
Given the news article and its reference summary, subtly alter every sentence of the summary to introduce EXACTLY ONE varied misrepresentations—such as incorrect entities, dates, or details without diverging drastically from the original structure.\\
Article:\{article\}\\
Summary:\{summary\}\\

\midrule
\end{tabular}

\begin{tabular}{p{15.4cm}l}
\\
\textbf{GPT-4 Extrinsic Hallucination}\\
\toprule
For each sentence in the provided summary of the news article, embed a distinctive, external detail not present in the original article. Every modified sentence should contain this additional information. Ensure these insertions are credible and do not clash with the article's facts.\\
Article:\{article\}\\
Summary:\{summary\}\\

\midrule
\end{tabular}

\begin{tabular}{p{15.4cm}l}
\\
\textbf{Stable Beluga 2 Summarization}\\
\toprule 
\#\#\# System:\\
You are StableBeluga, an AI that follows instructions extremely well. Help as much as you can. Reply only German.\\
\#\#\# User: Generate a summary in German for the following article. The summary should be around 2 to 3 sentences.\\
Article: \{article\}\\
\#\#\# Assistant:\\
\midrule
\end{tabular}

\begin{tabular}{p{15.4cm}l}
\\
\textbf{Llama 2 7b Summarization}\\
\toprule 
\#\#\# Instruction:
Generate a summary in German for the provided article. The summary should be around 2 to 3 sentences.\\
Article: \{article\}\\
\#\#\# Assistant:\\
\midrule
\end{tabular}

\begin{tabular}{p{15.4cm}l}
\\
\textbf{Llama 2 7b 20Minuten Fine-tuning}\\
\toprule 
\#\#\# Instruction:
Generate a summary in German for the provided article. The summary should be around 2 to 3 sentences.\\
Article: \{article\}\\
\#\#\# Assistant:\\
\{summary\}\\
\midrule
\end{tabular}

\caption{List of all prompts that we use to summarize the articles of the \absinth~dataset, generate intrinsic and extrinsic hallucinations with GPT-4, and fine-tune Llama 2 7b on \textit{20Minuten}.}
\label{tab:prompts_dataset}
\end{table*}

\begin{table*}
\small
\centering

\begin{tabular}{p{15.4cm}l}
\\
\textbf{LeoLM 13b \absinth~Fine-tuning}\\
\toprule 
\#\#\# Instruction:
Analyze whether the given sentence is faithful to the article. If the sentence solely conveys information that comes directly from the article, without any additions or omissions, respond with `Faithful'. If the sentence contains information that is in direct contradiction to the article, respond with `Intrinsic Hallucination'. If the sentence introduces information or details that are not explicitly mentioned in the article itself, respond with `Extrinsic Hallucination'.\\
Article: \{article\}\\
Sentence: \{sentence\}\\ 
\#\#\# Answer:\\
\{label\}\\
\midrule
\end{tabular}

\begin{tabular}{p{15.4cm}l}
\\
\textbf{LLM Inconsistency Detection}\\
\toprule 
Analyze whether the given sentence is faithful to the article. If the sentence solely conveys information that comes directly from the article, without any additions or omissions, respond with `Faithful'. If the sentence contains information that is in direct contradiction to the article, respond with `Intrinsic Hallucination'. If the sentence introduces information or details that are not explicitly mentioned in the article itself, respond with `Extrinsic Hallucination'.\\
Article: \{article\}\\
Sentence: \{sentence\}\\
Label:\\
\bottomrule
\end{tabular}
\caption{Prompts used on the inconsistency detection task with LLMs.}
\label{tab:prompts}
\end{table*}

\section{Technical Details}
\label{sec:appendix_technical}

We use the HuggingFace Trainer API to fine-tune all models for summarization and inconsitency detection with \absinth. We train the LLMs with 4-bit QLoRA~\citep{dettmers2023qlora} on an Nvidia A100-80GB GPU and the smaller language  models with default fine-tuning on an Nvidia 3090 GPU. We set the temperature to 0 during inference for all LLMs. 
 \subsection{Summary Generation Details}
 %- Huggingface Trainer API for finetuning, inference with greedy decoding 
 %- Finetuning details mbart and mlongt5 (learning rate, num epochs, optimizer, context lengths)
% - Instruction tuning details llama2-7b (prompt, learning rate, num epochs, optimizer, qlora params, context length
 %- gpt4 (which version, access through openai api, temperature
 %  - point to intrinsic and extrinsic prompts
 
We train mBART,\footnote{\href{https://huggingface.co/facebook/mbart-large-cc25}{facebook/mbart-large-cc25}} mLongT5,\footnote{\href{https://huggingface.co/agemagician/mlong-t5-tglobal-base}{agemagician/mlong-t5-tglobal-base}} and LLama 2 7b\footnote{\href{https://huggingface.co/NousResearch/Llama-2-7b-hf}{NousResearch/Llama-2-7b-hf}} on \textit{20Minuten} to generate summaries for the \absinth~dataset\textemdash see fine-tuning details in Table~\ref{finetuning_details}. During inference, we apply beam search and a beam size of 3 with mBART and mLongT5, and greedy decoding with Llama 2 7b. To generate summaries with GPT-4, we use OpenAI API\footnote{\url{https://platform.openai.com/}} and the \texttt{gpt-4-0613} snapshot from June 13th, 2023 with a context window of 8,192 tokens.
%During inference, we used beam search with a beam size of 3 for summary generation using mBART and mLongT5, while for Llama 2-7b, greedy decoding was utilized.
Lastly, we use the Lm-Eval-Harness framework~\cite{eval-harness} to generate zero-shot summaries with Stable Beluga 2 with a context window of 4,096.\footnote{\href{https://huggingface.co/stabilityai/StableBeluga2}{stabilityai/StableBeluga2}} Table~\ref{tab:prompts_dataset} lists the prompts used to generate the summaries.
 
 \subsection{Inconsistency Detection Details}
 % - Used lm-eval-harness for performing all evaluations, temp=0.0, greedy decoding,
 % - Describe base models and corresponding context lengths
 % - for fewshot we used stratified (one sample per label)
 % - for fine-tuning base-model, learning rate, num_epochs, optimizer, qlora params, prompt
 %mbert and xlm-roberta details
 We evaluate all LLMs using the Lm-Eval-Harness framework on the \absinth~test split. Specifically, we evaluate zero-shot and few-shot with the following model checkpoints from HuggingFace: Llama 2 7b,\footnote{\href{https://huggingface.co/NousResearch/Llama-2-7b-hf}{NousResearch/Llama-2-7b-hf}} Llama 2 13b,\footnote{\href{https://huggingface.co/NousResearch/Llama-2-13b-hf}{NousResearch/Llama-2-13b-hf}} LeoLM-Mistral 7b,\footnote{\href{https://huggingface.co/LeoLM/leo-mistral-hessianai-7b}{LeoLM/leo-mistral-hessianai-7b}} LeoLM 7b,\footnote{\href{https://huggingface.co/LeoLM/leo-hessianai-7b}{LeoLM/leo-hessianai-7b}} and LeoLM 13b.\footnote{\href{https://huggingface.co/LeoLM/leo-hessianai-13b}{LeoLM/leo-hessianai-13b}} In the few-shot setting, we randomly select 3 samples (i.e.\ one per label) from the training split and shuffle them. Finally, we further fine-tune mBERT,\footnote{\href{https://huggingface.co/google-bert/bert-base-multilingual-cased}{google-bert/bert-base-multilingual-cased}} XLM-RoBERTa,\footnote{\href{https://huggingface.co/FacebookAI/xlm-roberta-base}{FacebookAI/xlm-roberta-base}} and LeoLM 13b on the \absinth~ training split. Table~\ref{finetuning_details} and Table~\ref{tab:prompts} provide the fine-tuning details and the corresponding prompts, respectively.

\end{document}